\documentclass[lettersize,journal]{IEEEtran}
\usepackage{amsmath,amsfonts}
\usepackage{algorithmic}
\usepackage{algorithm}
\usepackage{array}
\usepackage[caption=false,font=normalsize,labelfont=sf,textfont=sf]{subfig}
\usepackage{textcomp}
\usepackage{stfloats}
\usepackage{url}
\usepackage{verbatim}
\usepackage{graphicx}
\usepackage{cite}
\usepackage[usenames,dvipsnames,svgnames,table]{xcolor}
\usepackage{xspace}
\usepackage{amsthm,amsmath,amssymb,lipsum}
\usepackage{mathrsfs}
\usepackage{booktabs}
\usepackage{multirow}
\usepackage{multicol}
\usepackage{bbding}
\usepackage{amsmath}
\usepackage[colorlinks,
            linkcolor=blue,
            anchorcolor=blue,
            citecolor=blue]{hyperref}
\usepackage{mathtools}
\def\onedot{\ifx\@let@token.\else.\null\fi\xspace}
\def\eg{\emph{e.g}\onedot} 
\def\eg{\emph{e.g}\onedot} 
\def\ie{\emph{i.e}\onedot}

\def\wrt{w.r.t\onedot} 
\def\etal{\emph{et al}\onedot}
\makeatother

\def\Vec#1{{\boldsymbol{#1}}}
\def\Mat#1{{\boldsymbol{#1}}}

\usepackage{amsthm}

\begin{document}


\title{On Distilling the Displacement Knowledge for Few-Shot Class-Incremental Learning}

\author{Pengfei~Fang$^\dagger$, Yongchun~Qin$^\dagger$, Hui~Xue$^*$,~\IEEEmembership{Member,~IEEE}
		\IEEEcompsocitemizethanks{\IEEEcompsocthanksitem  P. Fang, Y. Qin and H. Xue are with the School of Computer Science and Engineering, Southeast University, Nanjing 210096, China and the Key Laboratory of New Generation Artificial Intelligence Technology and Its Interdisciplinary Applications (Southeast University), Ministry of Education, Nanjing 211189, China.
			\protect
			E-mail: \{fangpengfei, ycqin,  hxue\}@seu.edu.cn
		}
        \thanks{$^\dagger$ Equally contribution} 
        \thanks{$^*$ Corresponding author}
}
\markboth{Journal of \LaTeX\ Class Files,~Vol.~14, No.~8, August~2021}%
{Shell \MakeLowercase{\textit{et al.}}: A Sample Article Using IEEEtran.cls for IEEE Journals}




\maketitle

\begin{abstract}

Few-shot Class-Incremental Learning (FSCIL) addresses the challenges of evolving data distributions and the difficulty of data acquisition in real-world scenarios. To counteract the catastrophic forgetting typically encountered in FSCIL, knowledge distillation is employed as a way to maintain the knowledge from learned data distribution. Recognizing the limitations of generating discriminative feature representations in a few-shot context, our approach incorporates structural information between samples into knowledge distillation. This structural information serves as a remedy for the low quality of features. Diverging from traditional structured distillation methods that compute sample similarity, we introduce the Displacement Knowledge Distillation (DKD) method. DKD utilizes displacement rather than similarity between samples, incorporating both distance and angular information to significantly enhance the information density retained through knowledge distillation. Observing performance disparities in feature distribution between base and novel classes, we propose the Dual Distillation Network (DDNet). This network applies traditional knowledge distillation to base classes and DKD to novel classes, challenging the conventional integration of novel classes with base classes. Additionally, we implement an instance-aware sample selector during inference to dynamically adjust dual branch weights, thereby leveraging the complementary strengths of each approach. Extensive testing on three benchmarks demonstrates that DDNet achieves state-of-the-art results. Moreover, through rigorous experimentation and comparison, we establish the robustness and general applicability of our proposed DKD method.

\end{abstract}


\section{Introduction} \label{sec:intro}






In recent years, deep learning has notably advanced the field of computer vision, largely due to pre-training on extensive datasets. However, in practical settings, data frequently occurs as streams, necessitating that artificial intelligence systems efficiently adapt to real-time data flows. Consequently, incremental learning—particularly class-incremental learning (CIL) \cite{iCaRL,CIL-castro2018end,CIL-hou2019learning}—has garnered growing interest. In these dynamic environments, neural networks face the challenge of continually adapting to new tasks while preserving knowledge from previous tasks. The interdependence of neural network parameters complicates the differentiation between new and existing knowledge, potentially leading to catastrophic forgetting, marked by a disproportionate focus on newly introduced categories. While current CIL methods have proven effective in scenarios rich in new data, real-world applications often grapple with challenges such as data collection, annotation costs, and privacy concerns. Therefore, few-shot class-incremental learning (FSCIL) \cite{TOPIC} has emerged as a critical research focus. A prime example of this is in facial recognition systems, where the ability to sequentially integrate and recognize new users, each contributing minimal photographic data, epitomizes an ideal FSCIL application scenario.

\begin{figure}[t]
\centering
    \includegraphics[width=90mm]{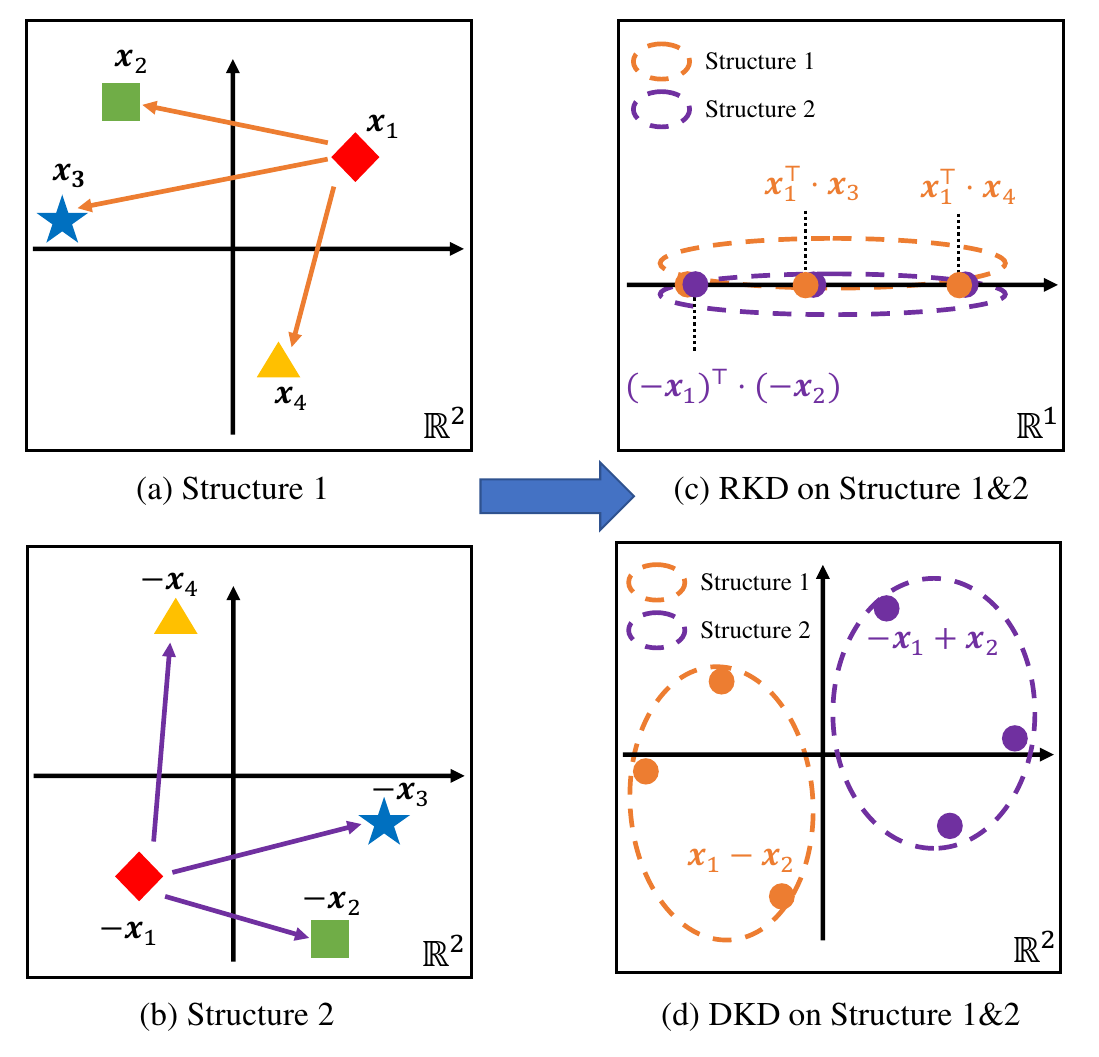}
    \caption{(a) and (b) respectively show two different structures with $\Vec{x}_1$ as the center, and are symmetric about the origin. (c) and (d) respectively show how RKD and DKD measure these two structures. The measure of RKD is 1-dimensional, and the obtained structure measure completely overlaps. DKD measures structural information in the original dimension so it can effectively distinguish between the two types of structures.}
     \label{fig:fig_1}
\end{figure}



Knowledge distillation stands out as one of the most frequently employed techniques in FSCIL, serving to transfer knowledge by preserving similarities between student and teacher outputs \cite{response-hinton2015distilling}. In CIL, the model from session $\tau$-1 is often utilized as the teacher model to oversee session $\tau$, thereby aiding memory retention during incremental learning. The majority of existing FSCIL methods rely on the logits of the networks for knowledge distillation, a process categorized as Individual Knowledge Distillation (IKD) according to \cite{RKD}, which focuses exclusively on the similarity between student-teacher pairs for individual samples. As highlighted by \cite{BiDist}, this approach may falter in the context of FSCIL, as the teacher network may struggle to acquire accurate feature representations due to insufficient data, thereby impeding the student network from assimilating knowledge from previous tasks during the knowledge distillation process.

In this case, Dong \etal \cite{RKD_FSCIL} propose that FSCIL can be facilitated through Relational Distillation Knowledge (RKD) \cite{RKD}. An Exemplar Relation Graph (ERG) is specifically constructed to establish ``teacher-student'' pairs, incorporating relational information among samples. This approach ensures the preservation of the topological structure during the knowledge distillation process. However, RKD faces two problems during distillation. Firstly, the coordinates of the samples in the feature space are weakened. As shown in Fig.~\ref{fig:fig_1}, as a low-dimensional metric, RKD may yield the same measure for samples with different structures, leading to imprecise structural modeling. Secondly, this relational distillation assesses the similarity between the central sample and nearby samples in each ``teacher-student" pair. Strengthening feature coupling between samples can undermine robustness against outliers, as the increased coupling may actually heighten sensitivity to anomalous data points. This issue is particularly critical in FSCIL, where limited samples are insufficient to correct biases introduced by outliers. To address these limitations, we introduce the \textbf{Displacement Knowledge Distillation} (DKD) approach, which offers a novel perspective on structural distillation within FSCIL. Specifically, given feature vectors of any two samples, DKD computes the displacement vector as their point-wise difference. Each dimension of this vector signifies the disparity between the samples along that dimension, subsequently transformed into a probability distribution through regularization. 
In DKD, each ``teacher-student" pair is distinct, preventing confusion, unlike RKD, where structural changes can occur without affecting individual metrics, DKD preserves structured modeling and avoids coupling features between samples, thus addressing RKD's robustness issues.

\begin{figure}
\centering
    \includegraphics[width=1.05\columnwidth]{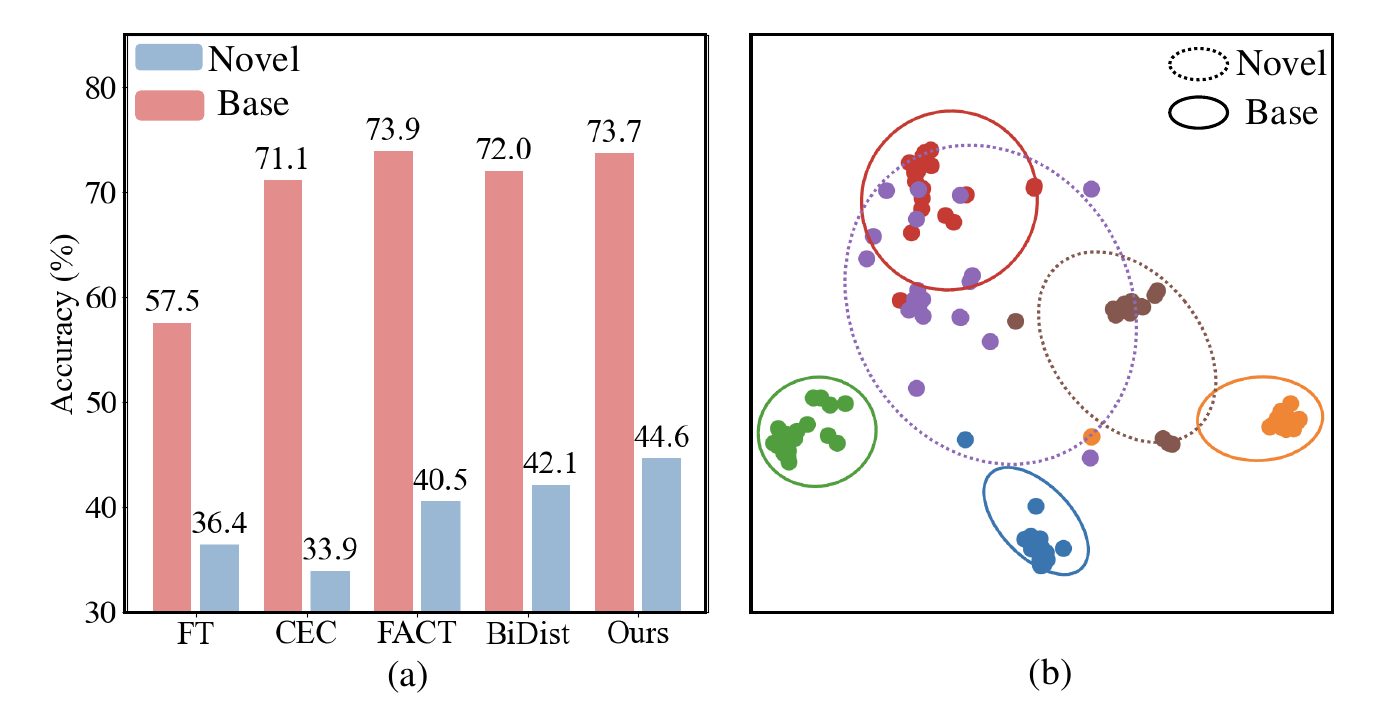}
    \caption{(a) Performance differences between base and novel classes on different methods. (b) The t-SNE visualization of features from base and novel classes respectively. According to the settings of FSCIL, the base classes have abundant training data and novel classes follow the few-shot setting, thus causing a performance gap between base and novel classes.}
     \label{fig:fig_3}
\end{figure}

A common approach in FSCIL is to pre-train the model on the base classes and then use the learned model to adapt to novel classes during the FSCIL stage, typically using the same distillation strategy. However, the significant gap between base and novel classes makes it challenging to learn discriminative features for the novel classes. As shown in Fig.~\ref{fig:fig_3}, there is a considerable performance gap between the base and novel features learned in FSCIL, which is evident in the tighter intra-class distribution of the base classes. Based on these observations, for the first time, we propose to divide base knowledge and new knowledge using two knowledge distillation strategies separately and design the \textbf{Dual Distillation Network (DDNet)} framework. 
On the one hand, the base classes are well-learned and can be represented by discriminative features, which makes it feasible to preserve knowledge directly through the output logits. On the other hand, distilling the output logits from only a few-shot set of categories poses challenges in learning effective features, as supported by empirical evidence. By preserving the structural relationships between samples, relational information can be utilized to improve knowledge retention\footnote{In the context of DKD, the novel classes refer to the samples from sessions $1$ to $\tau-1$, we name it pre-order dataset in Sec.~\ref{subsection:A} which is used for model distillation. In contrast, session $\tau$, which contains samples from the incremental classes, does not participate in the knowledge distillation phase.}. In doing so, we integrate a learnable, instance-aware sample selector into DDNet, allowing the model to distinguish between base and novel classes. By estimating the base and novel classes within this framework, DDNet can leverage knowledge from both distillation methods for class prediction during the inference phase. Our contributions include:


\begin{itemize}
    \item 
    We propose a \textbf{Displacement Knowledge Distillation (DKD)} method for structural knowledge preservation, which interprets the displacement between a sample pair as a probability distribution and preserves the similarity of this distribution between the teacher and student networks. This approach significantly enhances the retention of novel knowledge.
    \item 
    Based on observations of the impact of data scarcity in FSCIL, we observe that the traditional approach of merging novel classes into base classes is not suitable for FSCIL. Therefore, we design a \textbf{Dual Distillation Network (DDNet)} framework that preserves base knowledge through IKD and novel knowledge through DKD. In the inference stage, the knowledge from those two distillation methods is aggregated via the proposed sample selector.
    \item 
    We conducted extensive experiments on CIFAR-100, \textit{mini}ImageNet, and CUB-200, and fully compared our approach with existing methods. Our approach has demonstrated state-of-the-art performance across these three benchmark datasets, underscoring its efficacy and potential for practical application.
\end{itemize}
\section{Related Work} \label{sec:related}


\subsection{Few-shot Class-incremental Learning}
Few-shot learning (FSL) focuses on tackling machine learning challenges in environments where data availability is limited. The current FSL methods mainly include the meta-learning-based methods\cite{meta-ravi2016optimization,miniimagenet,meta-finn2017model,meta-snell2017prototypical,meta-rusu2018meta,meta-sung2018learning} and the metric-based methods\cite{metric-gidaris2018dynamic,metric-hou2019cross,metric-tian2020rethinking,metric-ye2020few,metric-ye2021learning,metric-yu2022hybrid}. Meta-learning provides good initialization parameters for FSL by learning on meta-tasks. Metric-based methods try to find the most appropriate distance metric for FSL. At the same time, increasing number of researches focus on open-world few-shot learning, such as noisy FSL\cite{noisyFSL-liang2022few}, cross-domain FSL\cite{CDFSL-chen2018closer}, class-incremental FSL\cite{TOPIC}, to name but a few, which lays a foundation for FSL to enter the practical application.

Class-incremental learning is a typical open-world task that requires models to continuously learn new tasks and retain knowledge of old tasks. Due to data collection constraints, labeling costs, and privacy concerns, few-shot class incremental learning (FSCIL) is crucial for training models with limited samples in real-world tasks~\cite{TOPIC}. The current FSCIL can be divided into three categories: data-based methods, ensemble-based methods and regularization-based methods.

\begin{itemize}
    \item \textbf{Data-based methods}. Such methods primarily focus on enhancing data utilization and typically require maintaining a memory set to store selected samples for data rehearsal. SPPR\cite{SPPR} designs a random episode selection strategy to obtain pseudo-samples and be applied in data rehearsal. S3C\cite{s3c} points out that data scarcity is an important cause of forgetting in FSCIL. 
    MoBoo~\cite{moboo} proposes a memory-augmented attention mechanism to achieve an expandable representation space within a fixed-size memory.
    In addition, some researches\cite{FeSSSS,UadCE,LCSL} propose a paradigm to assist FSCIL by introducing unlabeled data.
    \item \textbf{Ensemble-based methods}. Ensemble learning strives to integrate models with different properties and preferences. 
    BiDist\cite{BiDist} proposes a dual branch division method by base classes and novel classes, which fully take into account the over-fitting risk brought by inadequate training to base classes, so a fixed branch of base classes is adopted to retain knowledge.
    ALFSCIL~\cite{ALFSCIL} utilizes analogical learning to blend new and old classifiers and introduces the MAT module to ensure compatibility between new and old weights.
    MCNet\cite{MCNet} and BMC~\cite{BMC} combine two networks to alleviate catastrophic forgetting. Another type of ensemble-method usually uses multi-modal knowledge as guidance. For example, D'Alessandro \etal\cite{CPECLIP} take label texts as guidance information to assist feature embedding. Additionally, RBNL~\cite{RBNL} observes that the BN layer significantly impacts performance and achieves good results by removing it.
    \item \textbf{Regularization-based methods}.
    Identifying task-specific parameters within network parameters is a key concern of regularization. 
    WaRP\cite{WaRP} recognizes the parameters which are most important to the old knowledge by rotating the orthogonal space, thereby protecting the base knowledge by freezing those parameters. 
    From the perspective of mitigating forgetting, LFD~\cite{LFD} identifies the optimal optimization direction in the parameter space.
    TOPIC\cite{TOPIC} preserves the topology between prototypes through a neural gas network. Further, ERL\cite{RKD_FSCIL} defines the topology more precisely and raises it from class-level to instance-level. Zhang~\etal~\cite{CEC} and Wang~\etal~\cite{CEC+} study the impact of the relationships between different classifiers on the memory capacity of the model. FACT\cite{FACT} implements a forward-compatible incremental learning method by reserving virtual classes in advance. MICS\cite{MICS} demonstrates that leveraging base classes can effectively guide the learning of novel classes, while the soft labels derived from the Mixup method\cite{zhang2017mixup} offer robust guidance within the framework of FSCIL.
\end{itemize}

\subsection{Knowledge Distillation}

Knowledge distillation is a model compression and acceleration method in origin, but in recent years, it has been incorporated into many learning paradigms, enabling the capacity to mitigate forgetting. The essence of knowledge distillation is in the effective transfer of knowledge from the teacher network to the student network. Hinton \etal propose that students can learn from teacher network by imitation of teacher output\cite{response-hinton2015distilling}. According to \cite{gou2021knowledge-survey}, knowledge distillation can be divided into three parts for knowledge extraction. 1) In response-based knowledge\cite{response-ba2014deep,response-hinton2015distilling,response-kim2018paraphrasing,response-mirzadeh2020improved}, the output logit from the model is directly taken as supervisory information and the differences between students and teachers are measured by Kullback-Leibler divergence. But this approach often relies heavily on the classification head and lacks constraints on the middle layers of deep networks. 2) Feature-based knowledge\cite{feature-huang2017like,feature-komodakis2017paying,feature-ahn2019variational,feature-heo2019knowledge} extract the feature maps of the middle layers instead of the last layer to give the model cross-layer knowledge transfer capability. 3) Relation-based knowledge\cite{relation-yim2017gift,RKD,relation-lee2019graph,relation-liu2019knowledge,relation-tung2019similarity} not only consider the differences between student-teacher sample pairs, but also considers the structural relationship between samples. Our method follows the setting of relational distillation and proposes to model the structural relationship through displacement.

\section{Method} \label{sec:method}

\subsection{Problem Formulation}
\label{subsection:A}
In FSCIL, training set appears in the form of task sequence, and the data stream can be referred to as:$\{\mathcal{D}^{0},\mathcal{D}^{1},\cdots,\mathcal{D}^{\tau}\}$, where $\mathcal{D}^{\tau}=\{(x_i,y_i)\}$ is the training set for session $\tau$ and $y_i$ belongs to the class set $\mathcal{C}^{\tau}$. The training sets from different sessions do not share classes, and the test set in the $\tau$-th session contains all previous classes, \ie $\mathcal{C}^{0} \cap \mathcal{C}^{1}\cdots \cap \mathcal{C}^{\tau}=\emptyset$ and $C_{\mathrm{test}}=\mathcal{C}^{0} \cup \mathcal{C}^{1}\cdots \cup \mathcal{C}^{\tau}$. When learning the task in session $\tau$, the model can only access to $\mathcal{D}^{\tau}$ and $\mathbf{D}^{m}$, where $\mathbf{D}^{m}$ represents a memory set that stores instances from previous tasks for data replay. To clearly distinguish between base classes and novel classes, during the current session $\tau$, the pre-order dataset $\mathbf{D}^{p}$ is defined as $\mathbf{D}^{p}=\{\mathcal{D}^{1},\cdots,\mathcal{D}^{\tau-1}\}$. In such a method, different knowledge distillation operations need to be performed on instances from $\mathcal{D}^{0}$ and $\mathbf{D}^{p}$ respectively to alleviate catastrophic forgetting. Followed by the setting from \cite{TOPIC}, the model is trained with sufficient instances at the first session, \ie, $\tau = 0$. When $\tau > 0$, few-shot learning is performed according to the $N$-way $K$-shot setting, which means that each task contains only $N$ different categories, and each category has only $K$ samples.

\subsection{Network Overview}

\begin{figure*}
\centering
    \includegraphics[width=180mm]{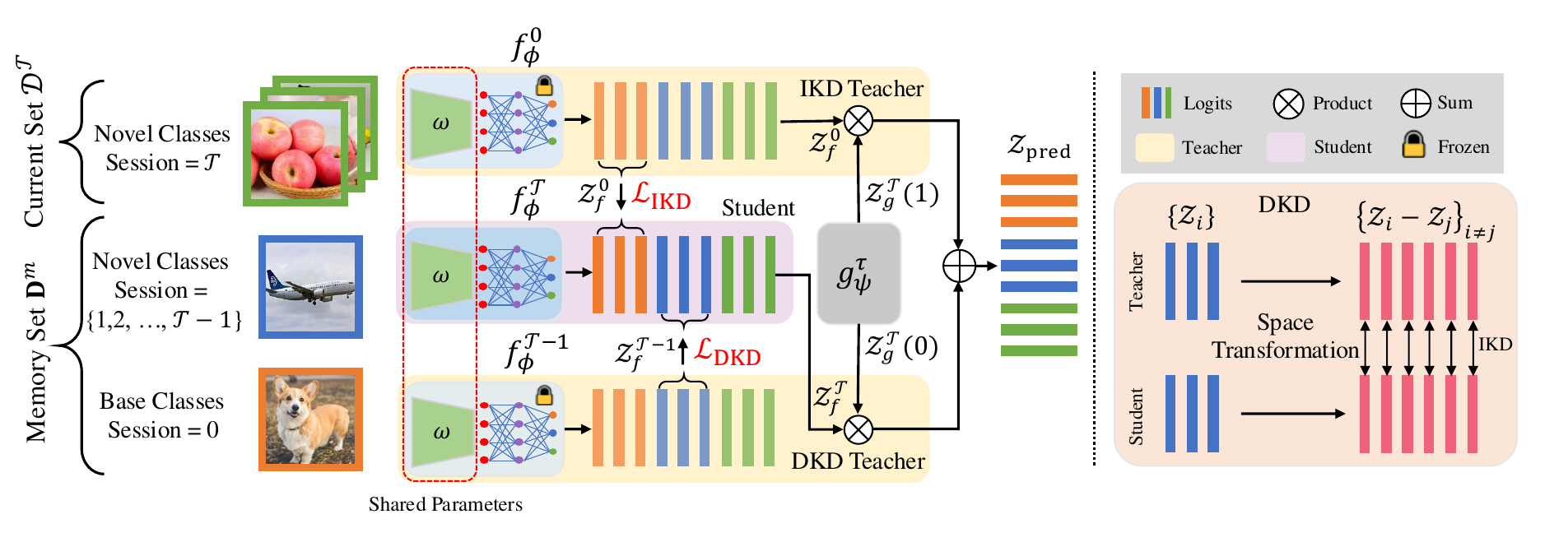}
    \caption{The framework of the DDNet and the illustration of DKD. We employs IKD to preserve the base knowledge and the proposed DKD method to protect the novel knowledge through the structural relationship. Logits from different sessions are integrated into a final prediction through the sample selector.}
     \label{fig:method}
\end{figure*}

We start by giving an overview of the proposed method. In common practice, FSCIL consists of two phases: standard supervised learning for base classes and class-incremental learning for novel classes. In the first phase, the model is trained with sufficient training sample from $\mathcal{D}^{0}$, thereby having a strong representation power for base classes. Given a sample $\Vec{x}$, the feature extractor of the model in session $0$ is defined as $f_{\phi}^{0}(\Vec{x})\in \mathbb{R}^{d}$, whose output is the $d$-dimensional feature vector. After the initial session, $f_{\phi}^{0}(\cdot)$ is frozen and saved as the base model.
In the second phase, a.k.a., incremental learning with a limited sample stage, the model from the previous session is fine-tuned using the current training set $\mathcal{D}^{\tau}$ and the memory set $\mathbf{D}^{m}$, forming $f_{\phi}^{\tau}(\cdot)$ in session $\tau$, where $\tau > 0$. 

The proposed DDNet employs IKD for the base classes and DKD, proposed in this paper, for the new classes. 
To determine whether a sample belongs to the base or novel classes, we design a sample selector, \( g_{\psi}^{\tau}(\Vec{x}) \in \mathbb{R}^{2} \), for a given sample \(\Vec{x}\). This selector enables DDNet to effectively differentiate between base and novel classes during testing.

Let the output logits of \( f_{\phi}^{\tau}(\cdot) \) and \( g_{\psi}^{\tau}(\cdot) \) be denoted as \(\Vec{z}_f^{\tau}\) and \(\Vec{z}_g^{\tau}\), respectively. The prediction \(\Vec{z}_{\mathrm{pred}}\) for a sample \(\Vec{x}\) is then obtained by weighting \(\Vec{z}_f^{\tau}\) and \(\Vec{z}_f^0\) according to the output \(\Vec{z}_{g}^{\tau}\) from the sample selector:

\begin{equation}\label{formula:pred_bin}
    \Vec{z}_{\mathrm{pred}}=\Vec{z}_{g}^{\tau}(0)\cdot \Vec{z}_f^{\tau}+\Vec{z}_{g}^{\tau}(1)\cdot \Vec{z}_f^{0}.
\end{equation}

Further details on the distillation process and the sample selector will be introduced in Sec.~\ref{subsection:C} and Sec.~\ref{subsection:D}.

\subsection{Knowledge Preserving in FSCIL}
\label{subsection:C}
Followed by \cite{RKD}, we divide knowledge distillation methods into three categories: IKD, RKD, and DKD, where IKD directly measures the logit while RKD and DKD measure the structural information. 
The key of our proposed DDNet is to select suitable knowledge distillation methods according to different feature distributions. Base classes are considered to have a superior feature representation because it has been fully and completely pre-trained. In order to avoid the catastrophic forgetting of the base knowledge caused by the overfitting of the novel few-shot classes, the best method is to maintain the output logits of base prediction directly. We take the logits $\Vec{z}^0_{f}$ as teacher and $\Vec{z}^{\tau}_{f}$ as student, then the IKD can be formulated as:
\begin{equation}\label{eq:ikd}
    \mathcal{L}_{\mathrm{IKD}}=\mathbb{E}_{(\Vec{x},y)\sim\mathbf{D}^{m}\cap\mathcal{D}^{0}} { KD}(\Vec{z}^{\tau}_{f}||\Vec{z}^0_{f}),
\end{equation}
where $KD(\cdot)$ represents the conventional knowledge distillation function which will be described in Sec.~\ref{sec:DPD}. The logits are obtained by the prototypical classification :
\begin{equation}
        \Vec{z}_f^{\tau}=\sigma \big({[{\Vec{p}_{1}^{\tau}},{\Vec{p}_{2}^{\tau}},\cdots,{\Vec{p}_{N}^{\tau}}]}^{\top}\cdot{f}_{\phi}^{\tau}(\Vec{x})\big),
\end{equation}
where $\sigma(\cdot)$ represents the softmax function. The class prototype is computed as the mean value of all sample features belonging to that class, given by the following equation:
\begin{equation}\label{equation:proto}
    \Vec{p}_{c}^{\tau}=\frac{1}{n_{c}}\sum_{\Vec{x}_{i}\in\mathcal{D}^{\tau},y_i=c}{f_{\phi}^{\tau}(\Vec{x}_{i})},
\end{equation}
where $n_{c}$ represents the number of instances of class $c$.
By employing IKD to retain the knowledge of base classes, the optimal features obtained by pre-training can be retained directly, which is conducive to preserving the classification performance of base classes. 

However, the empirical study \cite{IDLVQ} has suggested that the IKD is not an optimal solution to distill novel classes, as representation capabilities learned from few samples are weak. Instead, continuously adjusting the feature space for improved embedding is more effective, making it preferable to preserve instance relations rather than just retaining output logits.
\begin{figure*}[t]
\centering
    \includegraphics[width=180mm]{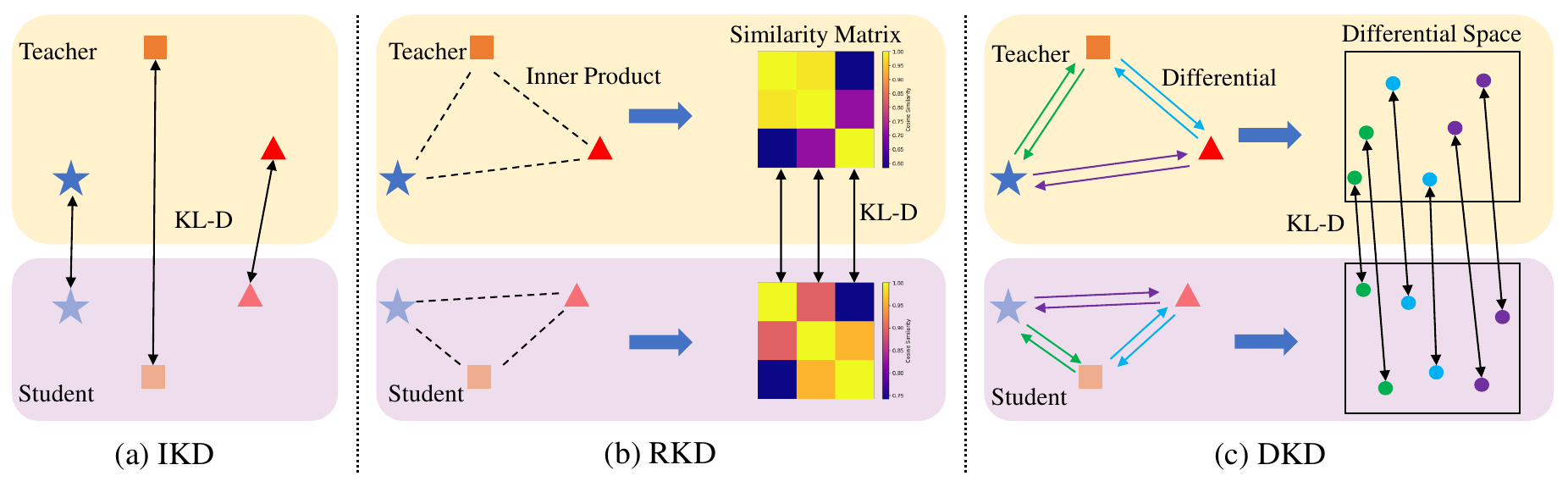}
    \caption{The illustration of differences between (a) IKD, (b) RKD, and (c) DKD. IKD directly computes the KL-divergence of teacher and student's output and thus cannot model the relationship between samples. RKD measures the relation of a sample pair via similarity, and the loss is the sum of KL-divergence of every row of the similarity matrix. DKD preserves all the structural information by making differences between samples and times the number of ``teacher-student'' pairs by $N-1$. Evidently, RKD leads to coupling between samples, whereas DKD completely avoids such relationships.}
     \label{fig:DKD}
\end{figure*}
As depicted in Fig.~\ref{fig:DKD}, the current research \eg RKD represents structures as a matrix where each row corresponds to a local structure centered on a specific sample. 
This approach establishes structural relationships, but also loses a significant amount of information, such as the direction relationships between samples. Consequently, this modeling approach is imprecise and poses challenges for the efficient retention of knowledge.


We propose that the displacement between any pair of samples can be utilized as guidance while distillation. By calculating the KL-divergence between the teacher and student distributions, the DKD method effectively retains knowledge of novel classes. The DKD loss is formulated as follows:
\begin{equation}\label{eq:dkd}
\begin{split}           
    \mathcal{L}_{\mathrm{DKD}}&=\mathbb{E}_{(\Vec{x},y)\sim\mathbf{D}^{m}\cap\mathbf{D}^{p}} {DKD}(\Vec{z}^{\tau}_{fi}||\Vec{z}^{\tau-1}_{fi}).
\end{split}
\end{equation}
\begin{equation}
\begin{split}
        {DKD}(&\Vec{z}^{\tau}_{fi}||\Vec{z}^{\tau-1}_{fi})= \\
        &\frac{1}{N-1} \sum_{j\neq i} {KD}\left({\Vec{z}^{\tau}_{fi}}-{\Vec{z}^{\tau}_{fj}} \| {\Vec{z}^{\tau-1}_{fi}}-{\Vec{z}^{\tau-1}_{fj}}\right),
\end{split}
\end{equation}
where $\Vec{z}^{\tau}_{fj}$ denotes the logit of sample $\Vec{x}_j$.


By differentiating the features, DKD eliminates interference from the actual sample values on structural relationships. The displacement vector fully retains all structural information between sample pairs, including both direction and distance, whereas RKD captures only one of these aspects. Additionally, since both the structure measure and sample features reside in \(\mathbb{R}^d\) space, there is a one-to-one correspondence between the real structure and the structure metric, avoiding ``misclassification'' of the structure. 
DKD treats each pair of samples as a distinct ``teacher-student'' pair, thereby effectively eliminating any coupling relationships between samples.

Through the combined use of IKD and DKD, the distillation loss function of our DDNet during knowledge retention is:
\begin{equation}
    \mathcal{L}_{\mathrm{kd}}=w_1\cdot\mathcal{L}_{\mathrm{IKD}}+w_2\cdot\mathcal{L}_{\mathrm{DKD}},
\end{equation}
in which $w_1$ and $w_2$ represent the weight of IKD and DKD loss, respectively.
The standard Cross-Entropy loss function is also used as the target:
\begin{equation}
\mathcal{L}_{\mathrm{cls}}=\mathbb{E}_{(\Vec{x},y)\sim\mathbf{D}^{m}\cup\mathcal{D}^{\tau}} {\rm CE}(\Vec{z}^{\tau}_{f},y).
\end{equation}

The loss function used to update the feature extractor $f_{\phi}^{\tau}(\cdot)$ is the sum of the classification loss and the distillation loss:

\begin{equation}
    \mathcal{L}_{f}=\mathcal{L}_{\mathrm{cls}}+\mathcal{L}_{\mathrm{kd}}.
\end{equation}

\subsection{Sample Selector}
\label{subsection:D}

The motivation of the sample selector is to mitigate the performance drop of base classes caused by catastrophic forgetting. Although knowledge distillation is adopted, forgetting is inevitable in incremental learning, but ${f}_{\phi}^{0}(\cdot)$ can retain all knowledge of base classes completely without subsequent incremental learning. Therefore, it is an effective approach to merge the outputs of ${f}_{\phi}^{\tau}(\cdot)$ and ${f}_{\phi}^{0}(\cdot)$. 
By strategically leveraging this performance divergence, our model effectively orchestrates the knowledge transfer, ensuring that the incremental learning system retains vital information from base classes and swiftly adapts to the nuances of the novel classes.

To adapt to the preferences of inputs by ${f}_{\phi}^{\tau}(\cdot)$ and ${f}_{\phi}^{0}(\cdot)$, the sample selector is designed to partition novel and base classes into two distinct clusters. The samples of the same class naturally share semantic similarity, however, samples in the novel or base cluster do not have this property, which is the key problem that makes the base-novel selector less effective.
We consider that the key to the problem is to specify a decision surface by introducing prior knowledge and forcing $g_{\psi}^{\tau}(\cdot)$ to fit the decision boundary.
By utilizing triplet loss to minimize the distance between the nearest pair of points in two clusters, novel and base classes can be effectively aggregated. This approach allows for the determination of an appropriate decision boundary by measuring the boundary distance between the two clusters.
In order to clearly define the optimization process, we formulate the internal structure of sample selector given a sample $\Vec{x}$:
\begin{equation}
\begin{split}
    g_{\psi}^{\tau}(\Vec{x}) &\coloneqq {[{\Vec{p}_{\mathrm{novel}}^{\tau}},{\Vec{p}_{\mathrm{base}}^{\tau}}]}^{\top}\cdot {w}_{\psi}^{\tau}(\Vec{x}),\\
\end{split}
\end{equation}
where ${w}_{\psi}^{\tau}(\cdot)$ outputs the $d$-dimensional feature vector and shares shallow parameters with $f_{\phi}^{\tau}(\cdot)$. Besides, ${\Vec{p}_{\mathrm{novel}}^{\tau}}$ and ${\Vec{p}_{\mathrm{base}}^{\tau}}$ represent prototypes for the novel and base cluster respectively, and is formulated by the momentum manner:
\begin{equation}
\begin{split}
    &\Vec{p}_{\mathrm{base}}^{\tau}\,=\alpha\frac{1}{n_{b}}\sum_{\Vec{x}_{i}\in\mathcal{D}^{\tau}\cup\mathbf{D}^{m}}^{y_i\in \mathcal{C}^{0}}{w}_{\psi}^{\tau}(\Vec{x}_{i})+(1-\alpha) \Vec{p}_{\mathrm{base}}^{\tau-1}\, , \\
    &\Vec{p}_{\mathrm{novel}}^{\tau}=\alpha\frac{1}{n_{n}}\sum_{\Vec{x}_{i}\in\mathcal{D}^{\tau}\cup\mathbf{D}^{m}}^{y_i\notin \mathcal{C}^{0}}{w}_{\psi}^{\tau}(\Vec{x}_{i})+(1-\alpha) \Vec{p}_{\mathrm{novel}}^{\tau-1},
\end{split}
\end{equation}
where $\alpha$ is the momentum weight which regulates the stability of the sample selector during incremental learning.

Then, the triplet loss is formulated as follows:
\begin{equation}
\begin{split}
\begin{split}
\mathcal{L}_{\mathrm{trip}}=\mathbb{E}_{(\Vec{x}_i,y_{i})\mathbf{D}^{m}\cup\mathcal{D}^{\tau}}&\sum_{j,k}^{y_j\in \mathcal{C}^{0},y_k\notin \mathcal{C}^{0}}\big[\Vert {w}_{\psi}^{\tau}(\Vec{x}_i)-{w}_{\psi}^{\tau}(\Vec{x}_j)\Vert_2 \\
    &-\Vert {w}_{\psi}^{\tau}(\Vec{x}_i)-{w}_{\psi}^{\tau}(\Vec{x}_k) \Vert_2 + {\gamma}\big],
\end{split}
\end{split}
\end{equation}
where $\gamma$ is the margin in triplet loss.


Due to the lack of natural semantic similarity among class clusters, prototypes may overlap. It is essential to incorporate Binary Cross-Entropy loss to enhance class differentiation:
\begin{equation}
\begin{split}    
    \mathcal{L}_{\mathrm{bincls}}&=\mathbb{E}_{(\Vec{x},y)\sim\mathcal{D}^{\tau}} {\rm CE}(\Vec{z}^{\tau}_{g},\mathbb{I}(y\in \mathcal{C}^{0})).\\
    \Vec{z}_g^{\tau}&=\sigma \big({g}_{\psi}^{\tau}(\Vec{x})\big),
\end{split}
\end{equation}

Therefore, the total loss function of the sample selector can be expressed as:
\begin{equation}
    \mathcal{L}_{g}=\beta_1\cdot\mathcal{L}_{\mathrm{trip}}+\beta_2\cdot\mathcal{L}_{\mathrm{bincls}},
\end{equation}
where $\beta_1$ and $\beta_2$ are flexible weight parameters of different loss functions. In the DDNet, the feature extractor $f_{\phi}^{\tau}$ and sample selector $g_{\psi}^{\tau}$ are optimized separately using the losses $\mathcal{L}_f$ and $\mathcal{L}_g$, respectively.

\section{Analysis on Displacement Distillation} \label{sec:DPD}


\subsection{Unified Definition on {IKD}, {RKD} and {DKD}}

This section delineates a unified framework for categorizing and defining diverse knowledge distillation paradigms. Suppose the output logits of the teacher and student network for samples in the same batch are given by:
\begin{equation}
    \Mat{Z}^s=\left[\Vec{z}_1^{s}, \Vec{z}_2^{s}, \ldots, \Vec{z}_N^{s}\right], 
    \Mat{Z}^t=\left[\Vec{z}_1^{t}, \Vec{z}_2^{t}, \ldots, \Vec{z}_N^{t}\right].
\end{equation}


IKD solely assesses the congruence between the student and teacher output logits for corresponding samples. For IKD, the knowledge distillation loss is formulated by the ``teacher-student'' pair, which can be formulated as:
\begin{equation}
    \mathcal{L}_{\mathrm{I K D}}\left(\Mat{Z}^s, \Mat{Z}^t\right)=\frac{1}{N} \sum_i K D\left({{\Vec{z}}_i^s} \| {{\Vec{z}}_i^t}\right),
\end{equation}
where $KD(\cdot)$ is the function of knowledge distillation. For a pair of ``teacher-student'' vectors $\Vec{v}^t$ and $\Vec{v}^s$, the $KD(\cdot)$ function usually adopts KL-divergence:
\begin{equation}
    KD\left(\Vec{v}^s \| \Vec{v}^t\right)=\sum_k \hat{v}_k^t \log \left(\frac{\hat{v}^s_k}{\hat{v}^t_k}\right),
\end{equation}
in which
\begin{equation}
    \hat{v}^s_k=\frac{\exp \left({v}^s_{k}\right)}{\sum_j\exp\left({v}^s_{j}\right)}, \hat{v}^t_k=\frac{\exp \left({v}^t_{k}\right)}{\sum_j\exp\left({v}^t_{j}\right)}.
\end{equation}

RKD computes the low-dimension relationship between samples. There are different forms in RKD and we define RKD as inner product here for the sake of discussion.\footnote{Euclidean distance and cosine similarity are used in the original RKD method\cite{RKD}. Similar conclusions can also be reached for other forms of RKD.} The definition of RKD is:
\begin{equation}
    \mathcal{L}_{\mathrm{R K D}}\left(\Mat{Z}^s, \Mat{Z}^t\right)=\frac{1}{N} \sum_i K D\left(\Vec{h}_i^s \| \Vec{h}_i^t\right),
\end{equation}
in which, $\Vec{h}^{s}_{i}=\sigma\left({{\Mat{Z}}^{s}}^{\top} \cdot {\Vec{z}}^{s}_{i}\right)$, $\Vec{h}^{t}_{i}=\sigma\left({{\Mat{Z}}^{t}}^{\top} \cdot {\Vec{z}}^{t}_i\right)$, representing the similarity between logits.

In contrast, DKD has a more accurate description of the relationship between samples. The definition of DKD is:
\begin{equation}
    \mathcal{L}_{\mathrm{DKD}}\left(\Mat{Z}^s, \Mat{Z}^t\right)=\frac{1}{N} \sum_i \frac{1}{N-1} \sum_{j\neq i} K D\left({\Vec{z}^s_i}-{\Vec{z}^s_j} \| {\Vec{z}^t_i}-{\Vec{z}^t_j}\right).
\end{equation}

The displacement relationship, calculated in DKD, as a point-wise information, enhances the information density during the distillation process.

    
    

\subsection{The Property of Displacement Distillation}

When analyzing DKD specifically, we compare and analyze different distillation methods by evaluating the gradients. For $KD\left(\Vec{v}^s \| \Vec{v}^t\right)$, the gradient is formulated as followed:
\begin{equation}\label{formula:gradient}
    \frac{\partial KD\left(\Vec{v}^s \| \Vec{v}^t\right)}{\partial\Vec{v}^s}=\sum_j \hat{v}_j^t \hat{\Vec{v}}^s-\hat{\Vec{v}}^t,
\end{equation}
where $\hat{v}_j^t$ is the $j$-th dimension in the vector $\hat{\Vec{v}}^t$.

According to Eq.~\eqref{formula:gradient}, the gradient of IKD can be formulated as:
\begin{equation}\label{formula:ikd_g}
    \nabla_{\Vec{z}_N^s} \mathcal{L}_{\mathrm{IKD}}\left(\Mat{Z}^s, \Mat{Z}^t\right)=\frac{1}{N}[\sum_k \hat{z}_{N k}^t \hat{\Vec{z}}_N^s-\hat{\Vec{z}}_N^t].
\end{equation}

In this context, we compute the gradient with respect to the last sample $\Vec{z}_N^s$ in the batch. Due to symmetry, similar conclusions apply to other samples. In Eq.~\eqref{formula:ikd_g}, $\hat{\Vec{z}}^t_N$ and $\hat{\Vec{z}}^s_N$ are vectors and $\hat{z}_{N k}^t$ is the $k$-th dimension of $\hat{\Vec{z}}^s_N$. We can find that the gradient direction for a sample $\Vec{z}_N^s$ is a linear combination of the outputs from teacher and student networks. 

\begin{figure}[h]
\centering
    \includegraphics[width=60mm]{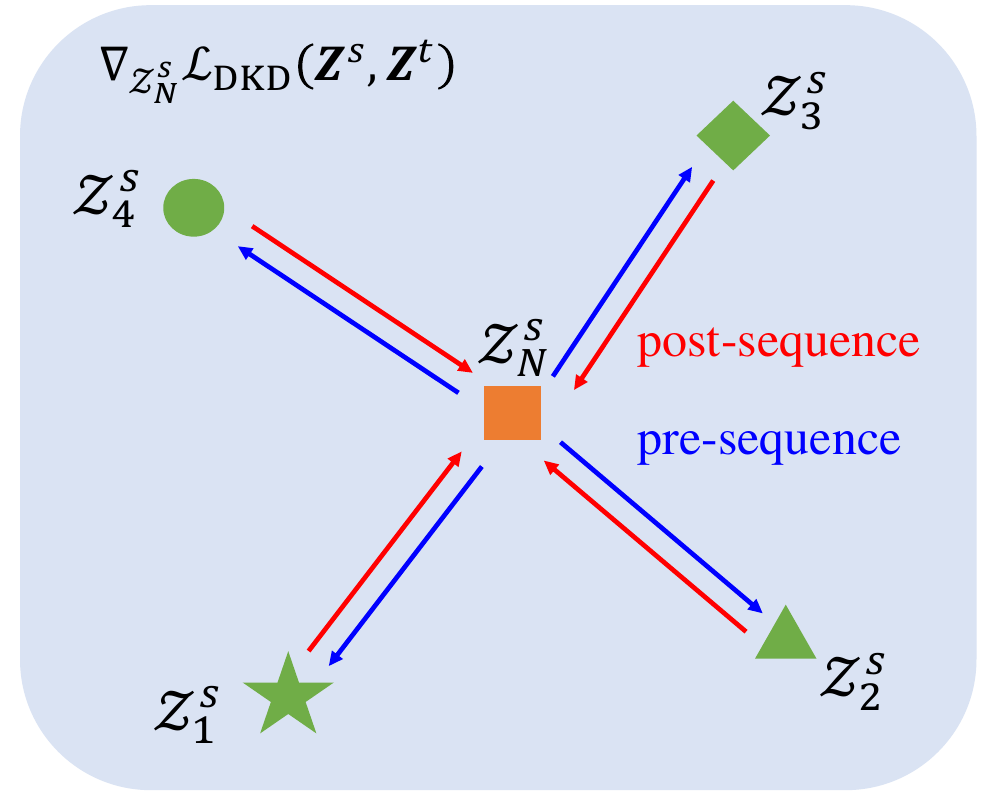}
    \caption{The illustration of the gradient of DKD. The red part represents the pre-sequence relation, and the blue is the post-sequence. Our proposed DKD includes bidirectional structural information of ``teacher-student'' pairs. }
     \label{fig:space_trans}
\end{figure}

Similarly, for DKD, the gradient with respect to $\Vec{z}_N^s$ can also be obtained, and the results are as follows:
\begin{equation}\label{formula:dpd_g1}
\begin{split}
    &\nabla_{\Vec{z}_N^s} \mathcal{L}_{\mathrm{DKD}}\left(\Mat{Z}^s, \Mat{Z}^t\right) \\
    =&\frac{1}{N (N-1)}\frac{\partial}{\partial{\Vec{q}^s_j}}\big[ {\color{red}\sum_{j\neq N} KD\left({\Vec{q}^s_j} \| {\Vec{q}^t_j}\right)} + {\color{blue}\sum_{j\neq N} KD\left({-\Vec{q}^s_j} \| {-\Vec{q}^t_j} \right)} \big],\\
\end{split}
\end{equation}
where we define ${\Vec{q}^{s(t)}_j}={\Vec{z}}_N^{s(t)}-{\Vec{z}}_j^{s(t)}$. The color in Eq.~\eqref{formula:dpd_g1} represents different structural relationships shown in Fig.~\ref{fig:space_trans}.

As DKD is a bidirectional measure of displacement, it has different gradient expressions when $\Vec{z}_N^{s(t)}$ is used as a pre-sequence node and a post-sequence node. As shown in the {\color{red}red} and {\color{blue}blue} parts of Eq.~\eqref{formula:dpd_g1} and Fig.~\ref{fig:space_trans}, when $\Vec{z}_N^{s(t)}$ serves as a pre-sequence node (\ie $\Vec{z}_N^{s(t)}-\Vec{z}_j^{s(t)}$), the gradient involves the displacement vector from $\Vec{z}_N^{s(t)}$ to all other samples, resulting in $N-1$ terms. When $\Vec{z}_N^{s(t)}$ is the post-order node (\ie $\Vec{z}_j^{s(t)}-\Vec{z}_N^{s(t)}$), there is only one displacement vector from each $\Vec{z}_j^{s(t)}$ node to $\Vec{z}_N^{s(t)}$, also resulting in $N-1$ term.

As Eq.~\eqref{formula:dpd_g1} shows, DKD can be formulated as the KL-divergence of $N(N-1)$ pairs of $\Vec{q}_j^{s(t)}$ vectors, in the sense that we create a set of vectors $\{\Vec{q}_j^{s(t)}\}_{j\neq N}$ by taking the difference of $\{\Vec{z}_N^{s(t)}-\Vec{z}_j^{s(t)}\}_{j\neq N}$. The property of DKD is actually a spatial transformation. The feature vector in the original space $\mathbb{R}^{d}$ is mapped to the difference space $\mathbb{D}^{d}$, and the coordinates of each sample in $\mathbb{D}^{d}$ no longer reflect the real information of the sample, but represent the relative relationship of a sample pair. Therefore, it can also be considered that the difference space $\mathbb{D}^{d}$ represents the structural information between samples in a mini-batch. 
From a spatial metric analysis perspective, RKD reduces the dimensionality by mapping samples from \(\mathbb{R}^d\) to \(\mathbb{R}^1\), resulting in substantial information loss. In contrast, DKD preserves the original dimensional information, preventing spatial information loss.

Compared to IKD and RKD, DKD involves more ``teacher-student" pairs. Both IKD and RKD only compute $N$ pairs, whereas DKD computes $N(N-1)$ pairs (considering the opposite direction as the same pair), thereby enhancing the efficiency of data utilization. 
In this regard, the proposed DKD can be understood as a way of data augmentation via the original data instead of the existing solution using pseudo-samples, which may interrupt the original data distribution.

\subsection{Robustness Analyses}

When compared with RKD, DKD avoids the high coupling between samples, thus improving the robustness. 
Building on the analyses above, we compute the gradient of RKD with respect to the last sample $\Vec{z}_N^s$ as follows:
\begin{equation}
\begin{split}
    &\nabla_{\Vec{z}_N^s} \mathcal{L}_{\mathrm{RKD}}\left(\Mat{Z}^s, \Mat{Z}^t\right)\\
    &=\frac{1}{N}\sum_j\left(\frac{\partial \Vec{h}_j^s}{\partial \Vec{z}_N^s}\right)^\top \cdot \nabla_{\Vec{h}_j^s} \mathcal{L}_{\mathrm{R K D}}\left(\Mat{Z}^s, \Mat{Z}^t\right) \\
    &=\frac{1}{N}\sum_j\left(\frac{\partial \Vec{h}_j^s}{\partial \Vec{z}_N^s}\right)^\top \cdot \nabla_{\Vec{h}_j^s} \mathcal{L}_{\mathrm{I K D}}\left(\Mat{H}^s, \Mat{H}^t\right),
\end{split}
\end{equation}
in which, $\Vec{h}_j^s$ is the $j$-th row in $\Mat{H}^s$:
\begin{equation}
    \Mat{H}^s=\left[\Vec{h}_1^{s\top}, \Vec{h}_2^{s\top}, \ldots, \Vec{h}_N^{s\top}\right].
\end{equation}

Further, the gradient can be formulated as follows:
\begin{equation}
    \nabla_{\Vec{h}_j^s} \mathcal{L}_{\mathrm{I K D}}\left(\Mat{H}^s, \Mat{H}^t\right)=\sum_k \hat{h}_{i k}^t \hat{\Vec{h}}_i^s-\hat{\Vec{h}}_i^t.
\end{equation}

We should note that the process of computing the similarity matrices $\Mat{H}^s$ and $\Mat{H}^t$ involves pairwise combinations of samples within the mini-batch. 
Therefore, the presence of an outlier can contaminate each term in the gradient of \(\mathcal{L}_{\mathrm{RKD}}\).


Suppose the outlier is \(\Vec{z}_N^s\), according to Eq.~\eqref{formula:dpd_g1}, there are \(N-1\) polluted terms in the gradient of DKD. The DKD gradient contains a total of \(N(N-1)\) terms, meaning that \(1/N\) of the "teacher-student" pairs are affected by the outlier during knowledge distillation, in contrast to 100\% in RKD. Therefore, in parallel computations involving multiple samples, our DKD method can significantly minimize the impact of outliers on the results.


Based on theoretical analyses, this kind of outlier error is an inherent defect of any distillation method based on similarity. Meanwhile, DKD can overcome this defect well, as displacement is an origin-to-dimension structural information. At the same time, we employ outliers to attack DKD in Sec.~\ref{sec:expt}, and the experimental result also shows that our method has even better robustness than current state-of-the-art methods.

\section{Experiments} \label{sec:expt}

\subsection{Datasets and Implementation Details}


\begin{table*}[htbp]
  \centering
  \caption{Comparisons to state-of-the-art FSCIL methods on CIFAR-100, \textit{mini}ImageNet, and CUB-200. Data comes from their paper except $^\dagger$ is conducted through the publicly available code from \cite{BiDist} in order to conduct a fair comparison. ``$M$'' represents the number of instances selected for data replay in each class. Bold texts: the best results, underline texts: the second-best results.}
    \begin{tabular}{lllllllllllllll}
    \toprule
    \multicolumn{1}{l}{\multirow{2}[2]{*}{Methods}} & \multicolumn{4}{c}{CIFAR-100} & \multicolumn{4}{c}{\textit{mini}ImageNet} & 
    \multicolumn{4}{c}{CUB-200}   & \multicolumn{2}{c}{Average} \\
          & $\mathrm{Acc}_0$  & $\mathrm{Acc}_\tau$  & KR~$\uparrow$    & AD~$\downarrow$    & $\mathrm{Acc}_0$  & $\mathrm{Acc}_\tau$  & KR~$\uparrow$    & AD~$\downarrow$    & $\mathrm{Acc}_0$  & $\mathrm{Acc}_\tau$  & KR~$\uparrow$    & AD~$\downarrow$    & KR~$\uparrow$    & AD~$\downarrow$    \\
    \midrule
    TOPIC\cite{TOPIC} & 64.10  & 15.85  & 24.73 & 48.25  & 61.31  & 24.42  & 39.83 & 36.89  & 68.68  & 26.26  & 38.24 & 42.42  & 34.26 & 42.52  \\
    EEIL~\cite{EEIL}  & 64.10  & 29.37  & 45.82 & 34.73  & 61.31  & 19.58  & 31.94 & 41.73  & 68.68  & 22.11  & 32.19 & 46.57  & 36.65 & 41.01  \\
    FACT~\cite{FACT}  & 78.83  & 51.84  & 65.76 & 26.99  & 75.32  & 48.51  & 64.41 & 26.81  & 78.91  & 55.96  & 70.92 & 22.95  & 67.03 & 25.58  \\
    CEC~\cite{CEC}   & 73.07  & 49.14  & 67.25 & 23.93  & 72.00  & 47.63  & 66.15 & 24.37  & 75.85  & 52.28  & 68.93 & 23.57  & 67.44 & 23.96  \\
    ERL++~\cite{RKD_FSCIL} & 73.63  & 48.32  & 65.63 & 25.31  & 61.79  & 40.88  & 66.16  & 20.91  & 73.52  & 52.28  & 71.11  & 21.24  & 67.63 & 22.49 \\
    MetaFSCIL~\cite{MetaFSCIL} & 74.50  & 49.97  & 67.07 & 24.53  & 72.04  & 49.19  & 68.28 & 22.85  & 75.90  & 52.64  & 69.35 & 23.26  & 68.24 & 23.55  \\
    ALFSCIL~\cite{ALFSCIL} & 80.75  & 55.17  & 68.32 & 25.58  & 81.27  & 53.31  & 65.60 & 27.96  & 79.79  & 59.30  & 74.32 & 20.49  & 69.41 & 24.68  \\
    F2M~\cite{F2M}   & 64.71  & 44.67  & \underline{69.03} & \textbf{20.04 } & 67.28  & 44.65  & 66.36 & 22.63  & 81.07  & 60.26  & 74.33 & 20.81  & 69.91 & \underline{21.16}  \\
    FCIL~\cite{FCIL}  & 77.12  & 52.02  & 67.45   &  25.10    &   76.34  &  52.76  & 69.11   & 23.58   &  78.70  &  58.48  &  74.31  & 20.22   & 70.29   & 22.97   \\
    WaRP~\cite{WaRP}  & 80.31  & 54.74  & 68.16 & 25.57  & 72.99  & 50.65  & 69.39 & 22.34  & 77.74  & 57.01  & 73.33 & 20.73  & 70.30 & 22.88  \\
    NC-FSCIL~\cite{NC-FSCIL} & 82.52  & 56.11  & 68.00 & 26.41  & 84.02 & 58.31  & 69.40 & 25.71  & 80.45  & 59.44  & 73.88 & 21.01  & 70.43 & 24.38  \\
    \midrule
    BiDist (M=1) $^\dagger$~\cite{BiDist} & 76.97  & 50.77  & 65.96 & 26.20  & 74.67  & 51.64  & 69.16 & 23.03  & 78.28  & 56.06  & 71.61 & 22.22  & 68.91 & 23.82  \\
    BiDist (M=5) $^\dagger$~\cite{BiDist} & 76.97  & 51.67  & 67.13 & 25.30  & 74.00  & 52.73  & \underline{71.26} & \underline{21.27}  & 78.28  & 57.02  & 72.84 & 21.26  & 70.41 & 22.61  \\
    \midrule
    Ours (M=1) & 76.97  & 52.21  & 67.83 & 24.76  & 74.00  & 52.22  & 70.57 & 21.78  & 78.28  & 58.94  & \underline{75.29} & \underline{19.34}  & \underline{71.23} & 21.96  \\
    Ours (M=5) & 76.97  & 53.41  & \textbf{69.39} & \underline{23.56}  & 74.00  & 53.95  & \textbf{72.91} & \textbf{20.05 } & 78.28  & 60.15  & \textbf{76.84} & \textbf{18.13 } & \textbf{73.05} & \textbf{20.58 } \\
    \bottomrule
    \end{tabular}%
  \label{tab:exps_all}%
\end{table*}%

\begin{figure*}[t]
\centering
    \includegraphics[width=0.99\linewidth]{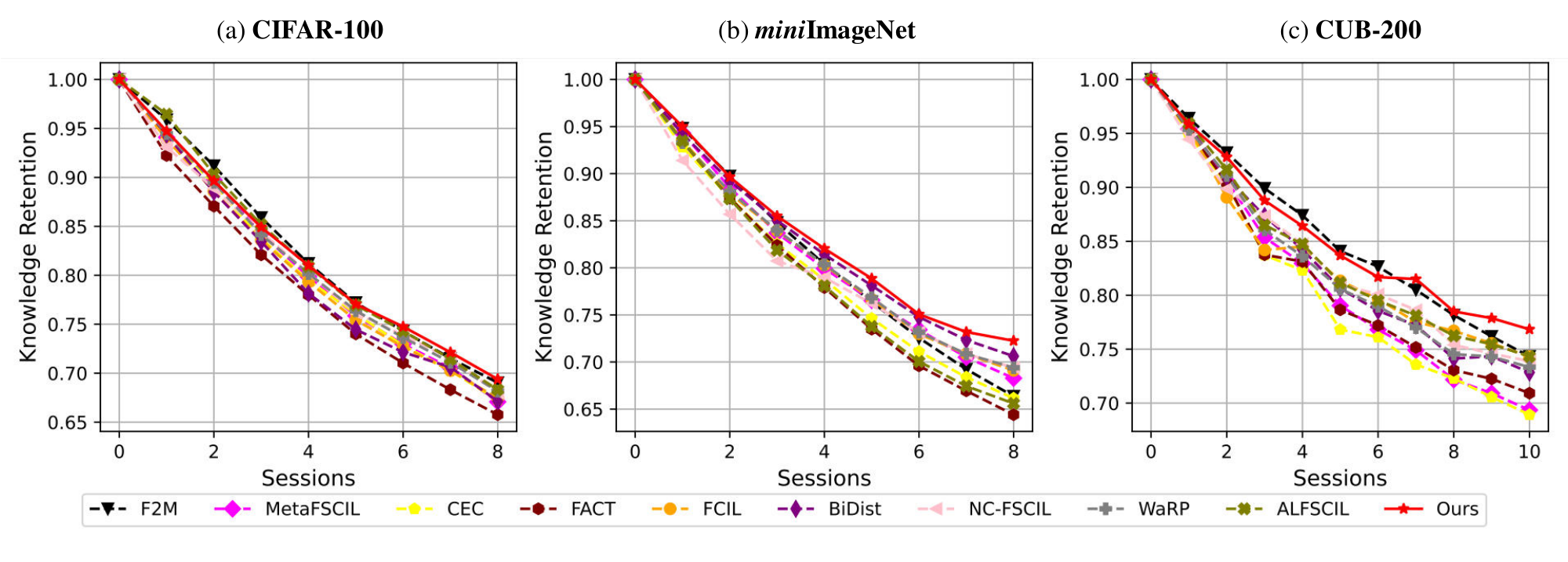}
    \caption{The Knowledge Retention rate of different methods on (a) CIFAR-100, (b) \textit{mini}ImageNet, and (c) CUB-200. Since a universal pre-trained model is not available for the FSCIL task, the initial state greatly influences the results. To ensure a fair comparison, we use the knowledge retention rate instead of average accuracy to assess the capabilities of different methods in incremental learning.}
     \label{fig:kd_compare}
\end{figure*}

According to the common experimental settings, we conduct experiments on CIFAR-100\cite{CIFAR100}, \textit{mini}ImageNet\cite{miniimagenet}, and CUB-200\cite{CUB} datasets. We follow the dataset split given by\cite{TOPIC} like most other methods to ensure a fair comparison. Besides, the backbone is kept the same as the compared methods. The details are listed below.
\begin{itemize}
    \item \noindent\textbf{CIFAR-100} contains 100 categories, each containing 500 training samples and 100 testing samples, and the size of each picture is 32 $\times$ 32. CIFAR-100 takes 60 classes as base classes (\ie, $\mathcal{D}^{0}$) and the remaining 40 classes are split by a $5$-way $5$-shot manner, resulting 8 incremental learning sessions ($\tau = 8$). 
    \item \noindent\textbf{\textit{mini}ImageNet} contains 100 categories sampled from ImageNet-1K\cite{imagenet1k}. Each category contains 500 training samples and 100 testing samples with a size of 84 $\times$ 84. Similar to CIFAR-100, \textit{mini}ImageNet adopts the same category split with 60 classes as base classes and 40 as novel in a $5$-way $5$-shot manner ($\tau = 8$).
    \item \noindent\textbf{CUB-200} is a fine-grained bird classification dataset containing 200 different bird species. For CUB-200, 100 classes are selected as base classes, and the remaining 100 classes are novel. The $10$-way $5$-shot manner is conducted, resulting 10 incremental learning sessions ($\tau = 10$).
\end{itemize}

Our approach is implemented by the Pytorch framework. We use ResNet-18\cite{resnet} for CUB-200 and \textit{mini}ImageNet, and ResNet-20\cite{resnet} for CIFAR-100 as the backbone network. We set the initial learning rate as 0.1 with cosine annealing, batch size as 64, training epoch as 200, and SGD as the optimizer on three datasets. For CIFAR-100 and \textit{mini}ImageNet, $\omega_1$ and $\omega_2$ are set as 50 identically and for CUB-200 as 30 and 50. $\beta_1$ and $\beta_2$ are set as 0.2 and 0.8, respectively. The momentum weight $\alpha$ equals 0.9 and the margin $\gamma$ in the triplet loss is specified as 1. Following \cite{BiDist}, for CUB-200, the model is initialized with the ImageNet-1K pre-trained parameters while for CIFAR-100 and \textit{mini}ImageNet the model is trained from scratch. Building upon insights gleaned from \cite{BiDist,MCNet}, the last residual block of the ResNet remains trainable during class-incremental learning, while the preceding blocks are frozen.
\subsection{Comparison with State-of-the-Art Methods}
We conduct comparison with other methods on three datasets. The compared methods cover three common categories of FSCIL, including TOPIC~\cite{TOPIC}, EEIL~\cite{EEIL}, FACT~\cite{FACT}, CEC~\cite{CEC}, ERL++~\cite{RKD}, MetaFSCIL~\cite{MetaFSCIL}, ALFSCIL~\cite{ALFSCIL}, F2M~\cite{F2M}, FCIL~\cite{FCIL}, WaRP~\cite{WaRP}, NC-FSCIL~\cite{NC-FSCIL}, and BiDist~\cite{BiDist}.
The results on \textit{mini}ImageNet, CIFAR-100, and CUB-200 are reported in Table~\ref{tab:exps_all}.
The test accuracy in session $\tau$ is denoted as Acc$_\tau$.
Because initial accuracy (Acc\(_0\)) significantly affects the final results (Acc\(_{\tau}\)), following~\cite{FCIL} ,we utilize Knowledge Retention rate (KR) and Accuracy Drop (AD) as the quantitative measure, defined as KR~=~Acc$_{\tau}$~/~Acc$_0 \times 100\%$ and AD~=~$\mathrm{Acc}_0-\mathrm{Acc}_{\tau}$.


As shown in Table~\ref{tab:exps_all}, the proposed DDNet outperforms the existing state-of-the-art methods across three benchmarks consistently \wrt the KR value. Specifically, our method achieves the final KR of 69.39\%, 72.91\%, and 76.84\% on the three dataset, and it brings the average performance gain of 2.62\%. At the same time, compared with other data replay methods, our method also has significant advantages, respectively leading FCIL ($M$=2), BiDist ($M$=1), and BiDist ($M$=5) by 0.94\%, 2.32\%, 2.64\%.

Through the analyses and summary of the data, we find that the methods based on model ensemble have comparative advantages. As the analyses from \cite{SPPR,BiDist,MCNet}, integrating neural networks with different performance profiles can effectively alleviate the stability-plasticity dilemma. Additionally, although methods based on feature regularization achieve significant results in the early stages, their recent performance has been lackluster. This decline is attributable to the highly coupled nature of neural networks, which complicates the task of isolating the roles of individual parameters. Consequently, determining the optimal direction for model optimization through regularization becomes exceedingly challenging. Meanwhile, more and more methods tend to combine different categories of FSCIL (\eg, FCIL, BiDist, and ours).




\begin{table*}[t]
  \centering
  \caption{Ablation studies of our proposed method on CIFAR-100. Acc$_b$ and Acc$_n$ are short for the accuracy of base and novel classes, respectively.  Bold texts: the best results.}
  \resizebox{\textwidth}{24mm}{
    \begin{tabular}{ccc|cc|ccccccccc|cc|cc}
    \toprule
    
    \multicolumn{3}{c|}{Feature Extractor} & \multicolumn{2}{c|}{Sample Selector} & \multicolumn{9}{c}{Accuracy in each session (\%)} & \multicolumn{2}{|c}{Session 8} & \multicolumn{2}{|c}{Overall} \\
    \midrule    
    IKD   & RKD   & DKD   & Momentum & triplet & 0     & 1     & 2     & 3     & 4     & 5     & 6     & 7     & 8     &   Acc$_{b}$    &  Acc$_{n}$ &KR$\uparrow$ &SA$\uparrow$ \\
    
    \midrule
    \Checkmark     &       &       &       &       & 76.97  & 72.62  & 67.83  & 64.03  & 60.63  & 57.65  & 56.31  & 53.39  & 51.61  & 70.98  & 22.55  & 62.34\%  & 61.74\%  \\
    \midrule
    \Checkmark     &       &       & \Checkmark     &       & 76.97  & 72.58  & 67.80  & 64.20  & 60.93  & 57.65  & 56.49  & 53.56  & 51.58  & 71.00  & 22.45  & 62.42\%  & 67.77\%  \\
    \Checkmark     &       &       &       & \Checkmark     & 76.97  & 72.71  & 68.06  & 64.36  & 60.86  & 57.61  & 56.43  & 53.58  & 51.53  & 70.88  & 22.50  & 62.46\%  & 64.23\%  \\
    \Checkmark     &       &       & \Checkmark     & \Checkmark     & 76.97  & 72.58  & 68.14  & 64.31  & 60.94  & 57.55  & 56.57  & 53.67  & 51.56  & 71.02  & 22.35  & 62.48\%  & \textbf{70.45}\%  \\
    \midrule
    \Checkmark     & \Checkmark     &       & \Checkmark     &       & 76.97  & 72.58  & 68.24  & 64.63  & 61.26  & 58.14  & 56.77  & 54.67  & 52.37  & 71.33  & 23.93  & 62.85\%  & 67.77\%  \\
    \Checkmark     & \Checkmark     &       &       & \Checkmark     & 76.97  & 72.71  & 68.67  & 64.67  & 61.48  & 58.09  & 56.84  & 54.58  & 52.32  & 71.40  & 23.70  & 62.93\%  & 64.23\%  \\
    \Checkmark     & \Checkmark     &       & \Checkmark     & \Checkmark     & 76.97  & 72.58  & 68.63  & 64.64  & 61.46  & 58.02  & 56.80  & 54.67  & 52.33  & \textbf{71.45}  & 23.65  & 62.90\%  & \textbf{70.45}\%  \\
    \midrule
    \Checkmark     &       & \Checkmark     & \Checkmark     &       & 76.97  & 72.58  & 68.33  & 64.84  & 61.96  & 58.84  & \textbf{57.59}  & \textbf{55.55}  & 53.60  & 71.20  & 27.20  & 63.36\%  & 67.77\%  \\
    \Checkmark     &       & \Checkmark     &       & \Checkmark     & 76.97  & 72.71  & 68.80  & 65.08  & 61.98  & 58.93  & 57.56  & 55.36  & 53.62  & 71.23  & 27.21  & 63.45\%  & 64.23\%  \\
    \Checkmark     &       & \Checkmark     & \Checkmark     & \Checkmark     & 76.97  & \textbf{72.91}  & \textbf{69.00}  & \textbf{65.35}  & \textbf{62.35}  & \textbf{59.31}  & 57.54  & 55.52  & \textbf{53.75}  & 71.37  & \textbf{27.31}  & \textbf{63.63}\%  & \textbf{70.45}\%  \\
    \bottomrule
    \end{tabular}}%
  \label{tab:addlabel}%
\end{table*}%

\begin{figure*}[t]
\centering
    \includegraphics[width=180mm]{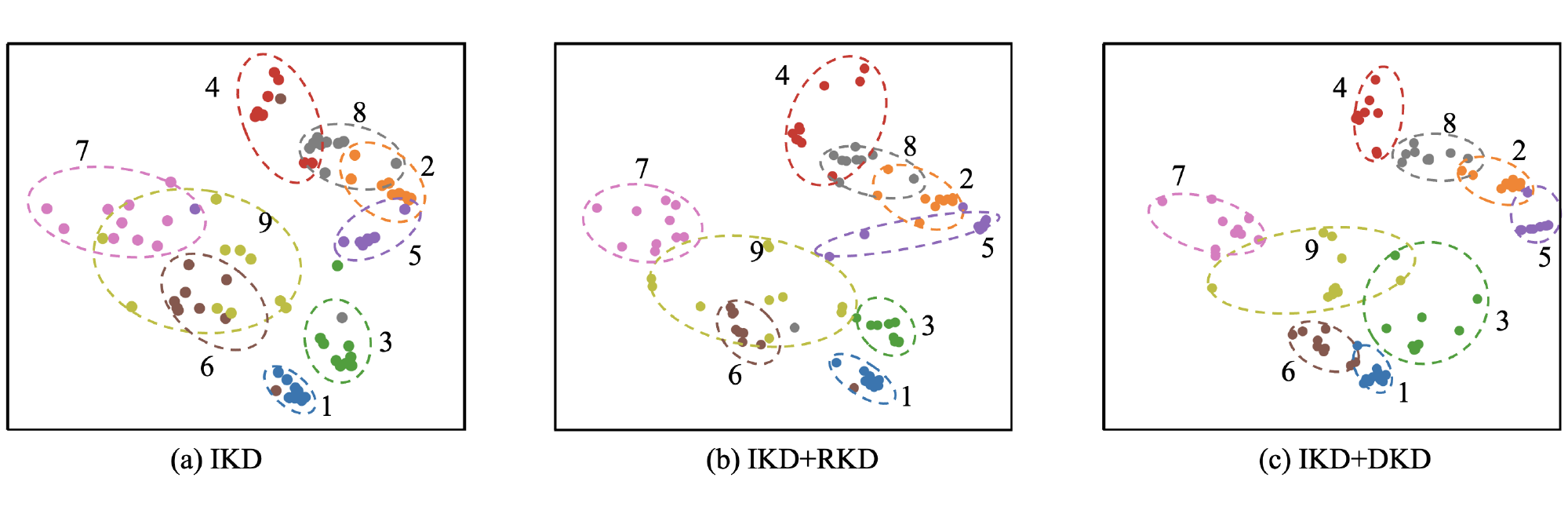}
    \caption{The t-SNE visualization of ablation study on CIFAR-100. Classes 1-5 are base classes, and classes 6-9 are novel classes. In (a), only IKD is applied in the knowledge distillation. In (b), IKD and RKD are employed, while IKD and DKD are adopted in (c). By comparing (a)(b)(c), we can find that both RKD and DKD help to separate different classes of samples in FSCIL, and DKD has a more pronounced effect.}
     \label{fig:ablation}
\end{figure*}

\begin{table*}[t]
  \centering
  \caption{Robustness studies of our proposed method on CIFAR-100. By attacking the learning process of the model by adding outliers, the performance drop of the model under different outliers is detected.}
    \begin{tabular}{lccccccccccc}
    \toprule
    \multirow{1.5}[2]{*}{Methods}&\multirow{1.5}[2]{*}{Outlier} & \multicolumn{9}{c}{Accuracy in each session (\%)}                       & \multirow{1.5}[2]{*}{AA$\uparrow$}\\
    \cmidrule{3-11} & & 0     & 1     & 2     & 3     & 4     & 5     & 6     & 7     & 8  &  \\

    \midrule
    BiDist & 0\%   & 76.97  & 66.09  & \textbf{63.09} & 59.69  & \textbf{57.80} & 53.96  & 53.40  & 51.71  & 49.17  & 59.10  \\
    Ours  & 0\%   & 76.97  & \textbf{70.11} & 60.50  & \textbf{60.41} & 56.94  & \textbf{54.45} & \textbf{53.81} & \textbf{51.79} & \textbf{50.29} & \textbf{59.47} \\
    \midrule
    BiDist & 1\%   & 76.97  & 65.15  & 61.46  & 59.36  & 56.27  & 53.78  & 52.43  & 50.55  & 48.93  & 58.32  \\
    Ours  & 1\%   & 76.97  & \textbf{69.49} & \textbf{61.61} & \textbf{59.97} & \textbf{57.59} & \textbf{53.86} & \textbf{52.99} & \textbf{51.73} & \textbf{49.92} & \textbf{59.35} \\
    \midrule
    BiDist & 5\%   & 76.97  & 65.82  & \textbf{62.77} & 58.57  & 56.16  & 53.46  & 51.82  & 51.06  & 48.56  & 58.35  \\
    Ours  & 5\%   & 76.97  & \textbf{69.60} & 61.21  & \textbf{59.31} & \textbf{56.34} & \textbf{54.29} & \textbf{53.53} & \textbf{51.76} & \textbf{49.28} & \textbf{59.14} \\
    \midrule
    BiDist & 10\%  & 76.97  & 65.35  & \textbf{62.61} & \textbf{59.43} & 56.12  & \textbf{54.35} & 51.64  & 50.39  & 48.24  & 58.34  \\
    Ours  & 10\%  & 76.97  & \textbf{69.58} & 59.90  & \textbf{59.43} & \textbf{56.88} & 54.20  & \textbf{53.12} & \textbf{51.38} & \textbf{48.55} & \textbf{58.89} \\
    \midrule
    BiDist & 15\%  & 76.97  & 63.77  & \textbf{62.43} & \textbf{59.25} & 56.17  & 52.54  & 51.27  & 50.08  & 47.55  & 57.78  \\
    Ours  & 15\%  & 76.97  & \textbf{69.22} & 60.60  & 58.56  & \textbf{57.05} & \textbf{54.36} & \textbf{52.72} & \textbf{51.42} & \textbf{49.40} & \textbf{58.92} \\
    \midrule
    BiDist & 20\%  & 76.97  & 63.91  & \textbf{61.39} & 58.29  & 56.04  & 53.13  & 51.89  & \textbf{50.36} & 47.84  & 57.76  \\
    Ours  & 20\%  & 76.97  & \textbf{69.54} & 59.50  & \textbf{58.52} & \textbf{56.46} & \textbf{53.85} & \textbf{52.23} & 50.35  & \textbf{48.45} & \textbf{58.43} \\
    \midrule
    \end{tabular}%

  \label{tab:robust}%
\end{table*}%

\subsection{Ablation Study}
Our ablation experiments mainly focus on the impact of different strategies on feature extractor $f_{\phi}^{\tau}(\cdot)$ and the sample selector $g_{\psi}^{\tau}(\cdot)$. 
First, we study the effect of different distillation strategies on our DDNet including IKD, RKD, and DKD.
Second, for the sample selectors, the contributions of momentum update and triplet loss are studied respectively. 
Besides KR, evaluation metrics applied here include Base Accuracy (Acc$_b$), Novel Accuracy (Acc$_n$), and Selector Accuracy (SA).



\noindent  {\bf The Impact of Different Distillation Strategies.} Regarding whether to use relational distillation, we carried out experiments in three settings (\ie, IKD, RKD, and DKD in Table~\ref{tab:addlabel}). Three settings represent the distillation method applied on samples from the pre-order dataset $\mathbf{D}^p$. 
RKD and DKD increase the Acc$_{b}$ by an average of 0.41\% and 0.30\%, respectively. However, relational distillation has a significant impact on Acc$_{n}$, with RKD and DKD increasing Acc$_{n}$ by an average of 1.33\% and 4.81\%, respectively. Two conclusions can be drawn. First, the relationship information between samples is vital in mitigating catastrophic forgetting. This relational information helps the model preserve the structural relationships between samples, leading to more accurate modeling of intra-class and inter-class relationships. Second, compared with the traditional RKD method, DKD method has significant advantages, because DKD transfers samples from the original space to the difference space, thereby multiplying the distillation pairs by $N-1$ times and preserving as much information as possible.

\noindent {\bf The Impact of Sample Selector.}  For sample selectors $g_{\psi}^{\tau}(\cdot)$, we test the validity of momentum update and triplet loss. As discussed in Sec.~\ref{sec:method}, a primary challenge of the sample selector is the clustering of categories that do not inherently share semantic similarity. Upon evaluation, the application of triplet loss alone results in a 2.49\% improvement in SA, while the use of momentum update alone yields a 6.03\% improvement. Combining both methods leads to a notable enhancement, with an overall improvement of 8.71\%. 
However, it is also observed that despite the increase in SA, the overall improvement in KR remains limited. The average increases for the three sample selector settings are only 0.53\%, 0.61\%, and 0.66\%, respectively. This phenomenon suggests that the network fusion based on Eq.~\eqref{formula:pred_bin} may benefit from further refinement, indicating significant opportunities for future research in this area.

\begin{figure}[t]
\centering
    \includegraphics[width=0.92\columnwidth]{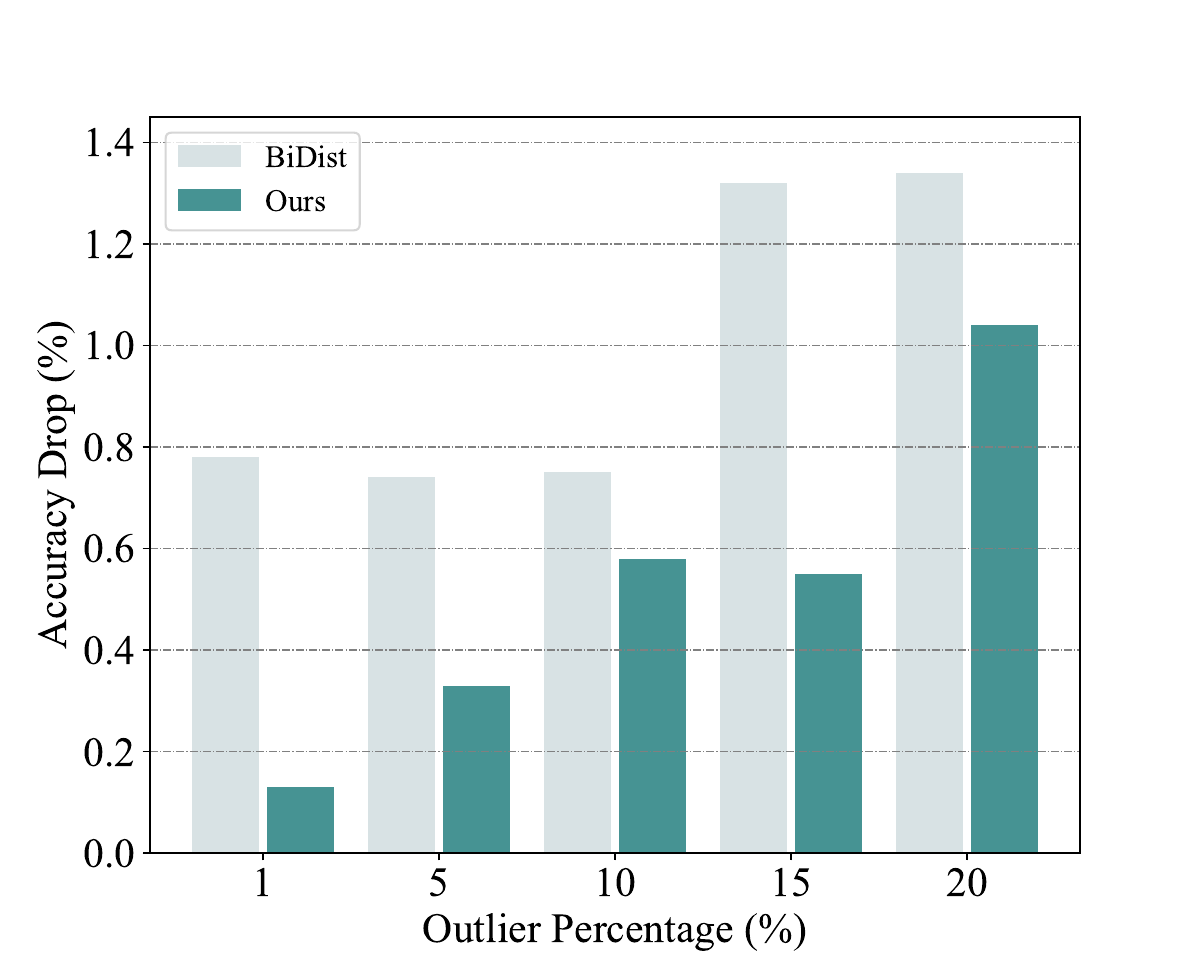}
    \caption{The impact of different outlier percentages on BiDist and our DDNet. Bold texts: the best results. Our method demonstrates significant robustness against outlier samples.}
     \label{fig:DPD_incre1}
\end{figure}

\subsection{Robustness of Displacement Distillation}
This part verifies the robustness of the proposed DKD method. The DDNet is attacked by randomly adding outliers to the training set $\mathcal{D}^{\tau}$ to observe the performance drop of the model in a noisy environment. By initializing with exactly the same pre-trained parameters, DDNet and BiDist can conduct a fair comparison. The metrics include Average Accuracy (AA) and Accuracy Drop (AD).

The proportion of outliers varies from 1\% to 20\%. Since both BiDist and our DDNet utilize a two-branch architecture, we focus solely on comparing the classification accuracy of the current branch (\ie, $f_{\phi}^{\tau}(\cdot)$ in DDNet) to more clearly demonstrate the impact of outliers. 
We report the results in Table~\ref{tab:robust}. DDNet outperforms the comparison method in various settings. Specifically, we can observe that our DKD-based method experiences a dramatic performance decline of approximately 10\% on average at $\tau$ = 2, precisely because DKD distillation of the pre-order dataset $\mathbf{D}^p$ begins when $\tau$ = 2. The semantic gap between base and novel classes results in performance degradation in early sessions. However, as FSCIL progresses, this semantic gap is gradually bridged, and our approach consistently outperforms the comparison methods in overall performance. Finally, on different outlier settings, our method reduces the averaged AD from 0.986\% to 0.526\% with a reduction of 46.7\% as shown in Fig.~\ref{fig:DPD_incre1}.

\subsection{Generalization of Displacement Distillation}

\begin{figure}[t]
\centering
    \includegraphics[width=1\linewidth]{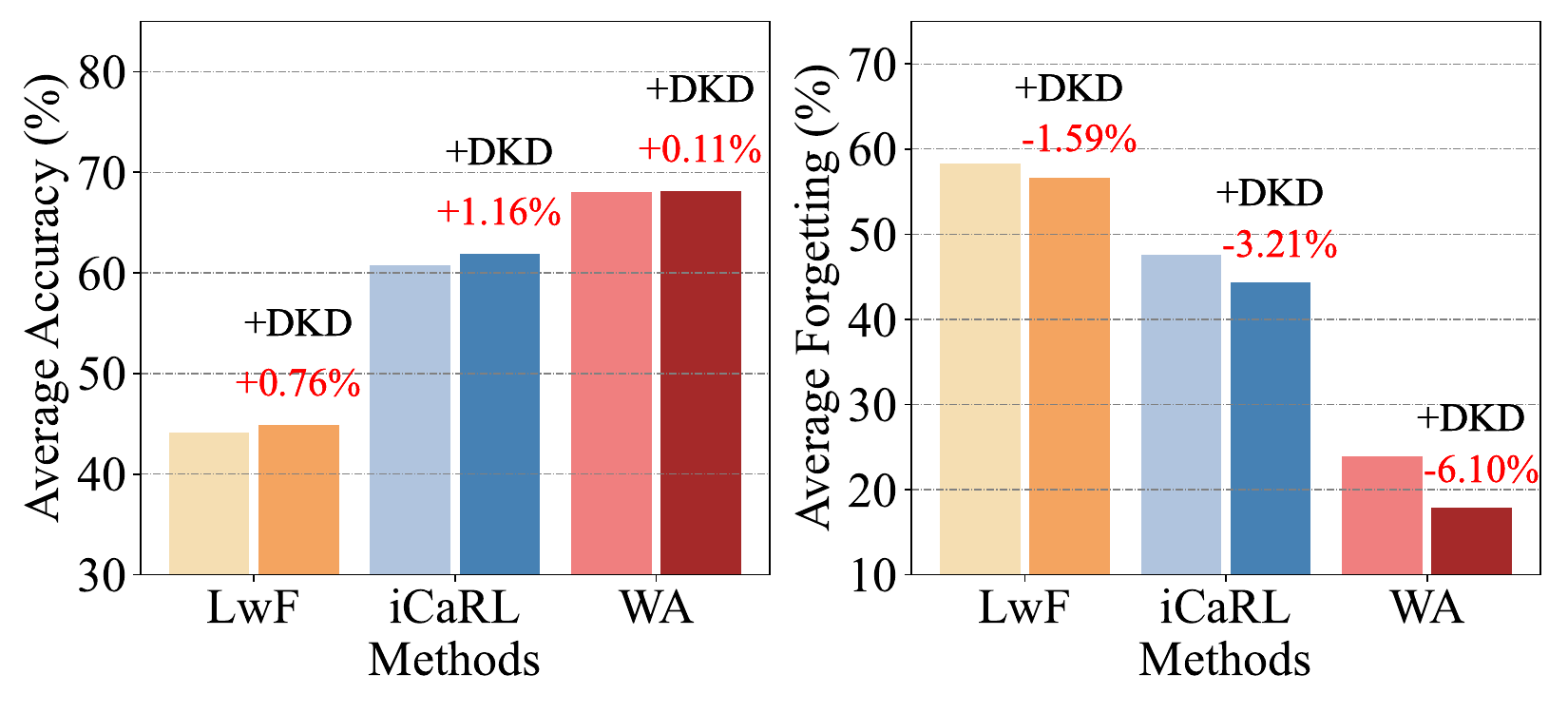}
    \caption{Studies about the effect of DKD on CIL. For the three methods, replacing the original IKD with DKD improves the average accuracy while mitigated the forgetting of previous knowledge.}
     \label{fig:DPD_incre2}
\end{figure}

To verify the generalization of our proposed DKD method, we carry out experiments on different tasks. For the more general CIL task, we select three representative CIL methods including LwF\cite{LwF}, iCaRL\cite{iCaRL}, and WA\cite{WA} and the results are presented in Fig.~\ref{fig:DPD_incre2}. In each of these methods, we replace IKD with DKD and observe that our method enhance model performance across all three. In this task, we adopt two evaluation metrics, AA and Average Forgetting (AF), proposed in \cite{forgetting}. AF measures the average forgetting rate of each session in incremental learning.

Our DKD method demonstrates remarkable improvements with compared methods. Experiments indicate that the DKD method is not only applicable to FSCIL but also can be applied to general CIL tasks. This versatility stems from DKD's ability to capture accurate structural information, aiding in the preservation of distributional consistency across different sessions in incremental learning.

\begin{table}[h]
  \centering
  \caption{Accuracy(\%) of different methods in knowledge distillation tasks on CIFAR-10. Experiments are conducted using the publicly available code from\cite{RKD}. Bold texts: the best results.}
    \begin{tabular}{llcccc}
    \toprule
    Teacher & Student & IKD   & RKD-D  & RKD-A & DKD \\
    \midrule
    ResNet-18\cite{resnet} & Conv-5 & 85.20  & 84.90 & 85.30  & \textbf{85.62} \\
    DenseNet-100\cite{desnet} & ResNet-18 & 87.30  & 88.10 & 88.40 & \textbf{89.27} \\
    PreResNet-110\cite{preresnet} & ResNet-18 & 87.20  & 88.20 & {89.30}& \textbf{89.90} \\
    ResNext-29-8\cite{resnext} & ResNet-18 & 84.40  & 86.00 & {88.30} & \textbf{88.40} \\
    WRN-28-10\cite{wrn}   & ResNet-18 & 84.60  & 86.70 & \textbf{87.10} & {86.70} \\
    \midrule
    \multicolumn{2}{c}{Average} & 85.74& 86.78& {87.68}& \textbf{87.98}\\
    \bottomrule
    \end{tabular}%
  \label{tab:DPD_dis}%
\end{table}%

In addition, we apply DKD to the model compression task, which is a common application of knowledge distillation. We select different teacher models and student models on the CIFAR-10 dataset for knowledge distillation. We compare the results of IKD, RKD-D, RKD-A, and DKD, where RKD-D represents the relational knowledge distillation with Euclidean distance while RKD-A represents the cosine similarity. The results are reported in Table~\ref{tab:DPD_dis}. DKD achieves the best results in four out of five different settings and outperforms other methods overall.
In comparison to the two RKD methods, DKD exhibits strong competitiveness in traditional knowledge distillation tasks, underscoring the importance of displacement information in knowledge preservation.

Through the application of DKD to both general class-incremental learning and knowledge distillation tasks, we conduct a comparative analyses of DKD-based methods against existing approaches. The experimental findings underscore that our DKD method serves as a versatile knowledge distillation technique, demonstrating potential applicability across diverse domains.

\section{Conclusion} \label{sec:conclusion}


This paper presents a novel strategy to tackle the challenge of catastrophic forgetting in few-shot class-incremental learning (FSCIL) via Displacement Knowledge Distillation (DKD). Our approach distills knowledge by emphasizing the structural relationships among samples, rather than focusing solely on individual output similarities. Additionally, we introduce the Dual Distillation Network (DDNet) framework, which seamlessly integrates Individual Knowledge Distillation (IKD) for base classes with DKD for newly introduced classes. Extensive experiments have confirmed that our method delivers state-of-the-art performance across diverse benchmarks, highlighting the effectiveness and broad applicability of the DKD approach. We believe our work will pave a new way to address this challenging task. Moving forward, we aim to further enhance the accuracy of the sample selector within DDNet to significantly boost FSCIL performance.


\bibliographystyle{ieeetr}
\bibliography{bib}

\begin{IEEEbiography}[{\includegraphics[width=1in,height=1.25in,clip,keepaspectratio]{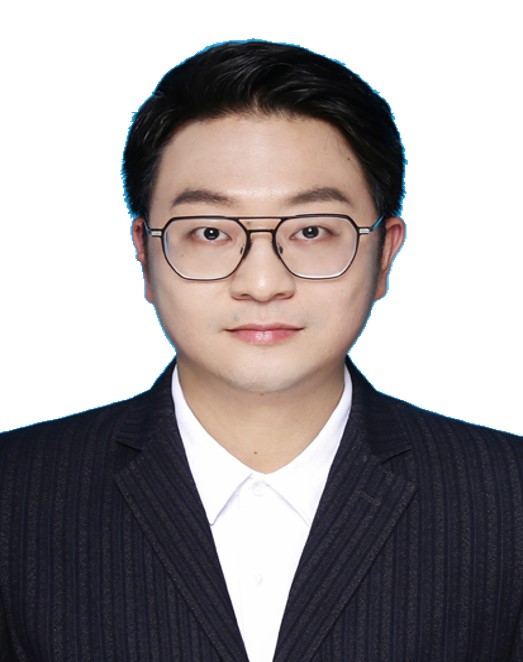}}]{Pengfei Fang} is an Associate Professor at the School of Computer Science and Engineering, Southeast University (SEU), China, and he is also a member of the PALM lab. Before joining SEU, he was a post-doctoral fellow at Monash University in 2022. He received the Ph.D. degree from the Australian National University and DATA61-CSIRO in 2022, and the M.E. degree from the Australian National University in 2017. His research interests include computer vision and machine learning. 
\end{IEEEbiography}

\vspace{-10cm}

\begin{IEEEbiography}[{\includegraphics[width=1.2in,height=1.4in,clip,keepaspectratio]{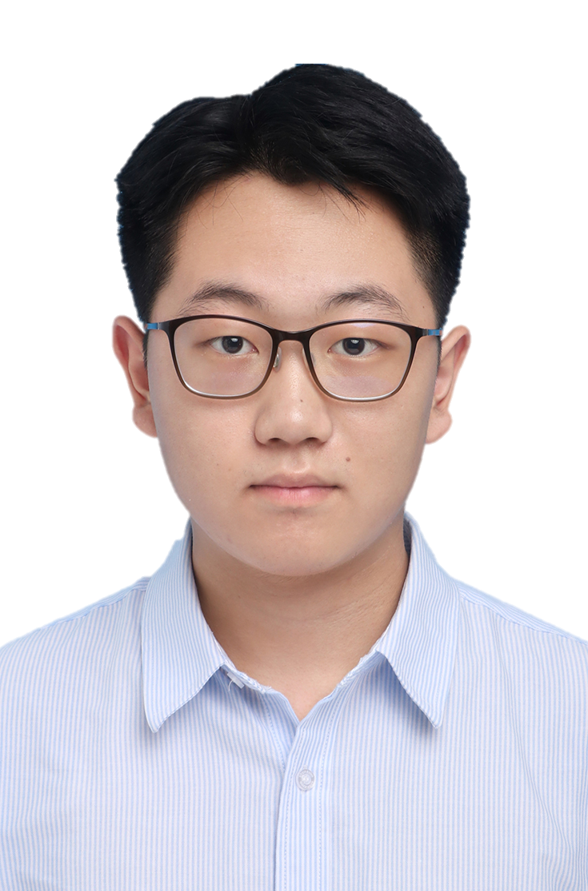}}]{Yongchun Qin} received the B.Sc. in artificial intelligence from Southeast University in 2023. He is currently pursuing the M.Sc. degree in School of computer science and engineering, at Southeast University. His research interest includes machine learning and pattern recognition.
\end{IEEEbiography}

\vspace{-10cm}

\begin{IEEEbiography}[{\includegraphics[width=1in,height=1.25in,clip,keepaspectratio]{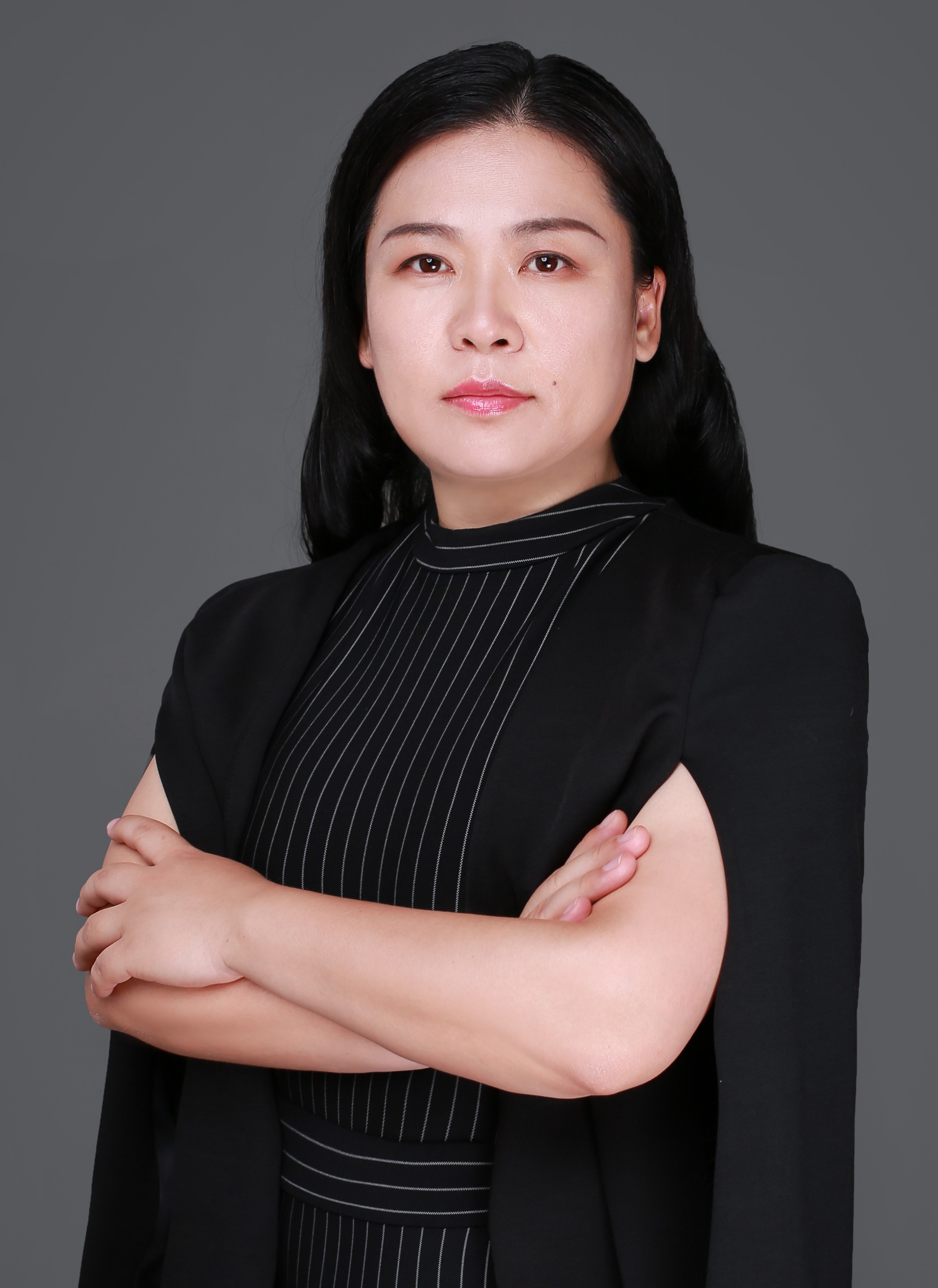}}]{Hui Xue}  received the B.Sc. degree in mathematics from Nanjing Norm University, Nanjing, China, in 2002, and the M.Sc. degree in mathematics and the Ph.D. degree in computer application technology from Nanjing University of Aeronautics and Astronautics (NUAA), Nanjing, in 2005 and 2008, respectively. She is currently a Professor with the School of Computer Science and Engineering, Southeast University, Nanjing, China. Her research interests include pattern recognition and machine learning
\end{IEEEbiography}

\end{document}